\documentclass[11pt,a4paper]{article}
\usepackage[nohyperref]{acl2017}
\usepackage{times}
\usepackage{latexsym}
\usepackage{amsmath}
\usepackage{amsfonts}
\usepackage{url}
\usepackage{graphicx}
\usepackage{booktabs}
\usepackage{adjustbox}
\usepackage{subfigure}
\usepackage{epstopdf}
\usepackage{verbatim}

\aclfinalcopy 
\def\aclpaperid{211} 

\title{A Hybrid Convolutional Latent Variable Generative Model of Natural Text}
\title{A Hybrid Convolutional Variational Autoencoder for Text Generation}

\author{Stanislau Semeniuta$^{1}$ \; Aliaksei Severyn$^{2}$ \; Erhardt Barth$^{1}$\\
$^{1}$Universit{\"a}t zu L{\"u}beck, Institut f{\"u}r Neuro- und Bioinformatik \\
 \normalsize{\tt \{stas,barth\}@inb.uni-luebeck.de}\\
$^{2}$Google Research \\
 \normalsize{\tt severyn@google.com}\\
}
\date{}

\begin{document}
\maketitle

\begin{abstract}
In this paper we explore the effect of architectural choices on learning a Variational Autoencoder (VAE) for text generation.
In contrast to the previously introduced VAE model for text where both the encoder and decoder are RNNs, we propose a novel hybrid architecture that blends fully feed-forward convolutional and deconvolutional components with a recurrent language model. Our architecture exhibits several attractive properties such as faster run time and convergence, ability to better handle long sequences and, more importantly, it helps to avoid some of the major difficulties posed by training VAE models on textual data.
\end{abstract}


\section{Introduction}

Generative models of texts are currently at the cornerstone of natural language understanding enabling recent breakthroughs in machine translation~\cite{DBLP:journals/corr/BahdanauCB14, DBLP:journals/corr/WuSCLNMKCGMKSJL16}, dialogue modelling~\cite{DBLP:journals/corr/SerbanSLCPCB16}, abstractive summarization~\cite{DBLP:journals/corr/RushCW15}, etc.

Currently, RNN-based generative models hold state of the art results in both unconditional \cite{DBLP:journals/corr/JozefowiczVSSW16, DBLP:journals/corr/HaDL16} and conditional \cite{DBLP:journals/corr/VinyalsTBE14} text generation. At a high level, these models represent a class of autoregressive models that work by generating outputs sequentially one step at a time where the next predicted element is conditioned on the history of elements generated thus far.
\begin{table}
	\footnotesize
	\begin{tabular}{l}
		\toprule
		@userid @userid @userid @userid @userid             \\
		@userid thanks for the follow                       \\
		@userid @userid @userid @userid @userid             \\
		@userid @userid @userid @userid @userid             \\
		@userid thanks for the follow                       \\ \midrule
		@userid All the best!!                              \\
		@userid you should come to my house tomorrow        \\
		I wanna go to the gym and I want to go to the beach \\
		@userid and it's a great place                      \\
		@userid  I hope you're feeling better               \\ \bottomrule
	\end{tabular}
	\caption{Examples of randomly generated tweets by a VAE model with a close to zero (top section) and larger than zero (bottom) KL term values. 
	}
	\label{table:kl_samples}
	\vspace{-1.5em}
\end{table}

Variational Autoencoders (VAE), recently introduced by \cite{vae_kingma, vae_rezende}, offer a different approach to generative modeling by integrating stochastic latent variables into the conventional autoencoder architecture. 
The primary purpose of learning VAE-based generative models is to be able to generate realistic examples as if they were drawn from the input data distribution by simply feeding noise vectors through the decoder. Additionally, the latent representations obtained by applying the encoder to input examples give a fine-grained control over the generation process that is harder to achieve with more conventional autoregressive models. Similar to compelling examples from image generation, where it is possible to condition generated human faces on various attributes such as hair, skin color and style~\cite{DBLP:journals/corr/YanYSL15,DBLP:journals/corr/LarsenSW15}, in text generation it should be possible to also control various attributes of the generated sentences, such as, for example, sentiment or writing style.

While training VAE-based models seems to pose little difficulty when applied to the tasks of generating natural images \cite{NIPS2016_6141,PixelVAE:DBLP:journals/corr/GulrajaniKATVVC16} and speech \cite{NIPS2016_6039}, their application to natural text generation requires additional care~\cite{generating_sentences, DBLP:journals/corr/MiaoYB15}. As discussed by \citet{generating_sentences}, the core difficulty of training VAE models is the collapse of the latent loss (represented by the KL divergence term) to zero. In this case the generator tends to completely ignore latent representations and reduces to a standard language model. This is largely due to the high modeling power of the RNN-based decoders, which with sufficiently small history can achieve low reconstruction errors while not relying on the latent vector provided by the encoder. Table \ref{table:kl_samples} shows that a VAE model where the KL term collapses to zero generates repeating and uninteresting samples.

In this paper, we propose a novel VAE model for texts that is more effective at forcing the decoder to make use of latent vectors. Contrary to existing work, where both encoder and decoder layers are LSTMs, the core of our model is a feed-forward architecture composed of one-dimensional convolutional and deconvolutional \cite{DBLP:conf/cvpr/ZeilerKTF10} layers. This choice of architecture helps to gain more control over the KL term, which is crucial for training a VAE model. Given the difficulty of generating long sequences in a fully feed-forward manner, we augment our network with an RNN language model layer. 
To the best of our knowledge, this paper is the first work that successfully applies deconvolutions in the decoder of a latent variable generative model of natural text. We empirically verify that our model is easier to train than its fully recurrent alternative, which, in our experiments, fails to converge on longer texts. To better understand why training VAEs for texts is difficult we carry out detailed experiments, discuss optimization difficulties, and propose effective ways to address them. Finally, we demonstrate that sampling from our model yields realistic texts. 


\section{Related Work}

Currently, there are three major streams of approaches to generative modeling: the Variational Autoencoder \cite{vae_kingma, vae_rezende}, autoregressive models \cite{DBLP:journals/jmlr/LarochelleM11, DBLP:journals/corr/OordKK16} and Generative Adversarial Networks (GAN)  \cite{DBLP:journals/corr/GoodfellowPMXWOCB14}.

Autoregressive models are built on the assumption that the current data element can be accurately predicted given sufficient history of elements generated thus far. 
The conventional RNN based language models fall into this category and currently dominate the language modeling and generation problem in NLP. Neural architectures based on recurrent \cite{DBLP:journals/corr/JozefowiczVSSW16, DBLP:journals/corr/ZophL16, DBLP:journals/corr/HaDL16} or convolutional decoders \cite{DBLP:journals/corr/KalchbrennerESO16, DBLP:journals/corr/DauphinFAG16} provide an effective solution to this problem. 

GANs have proven to be very effective in the Computer Vision domain \cite{DBLP:journals/corr/RadfordMC15, DBLP:journals/corr/DentonCSF15, DBLP:journals/corr/SalimansGZCRC16} but so far have gained little traction in the NLP community~\cite{DBLP:journals/corr/YuZWY16, text_gan}. 

\citet{generating_sentences} is a recent work that tackles language generation problem within the VAE framework. The authors demonstrate that with some care it is possible to successfully learn a latent variable generative model of text. Although their model is slightly outperformed by a traditional LSTM~\cite{Hochreiter:1997:LSM:1246443.1246450} language model, their model achieves a similar effect as in computer vision where one can (i) sample realistic sentences by feeding randomly generated novel latent vectors through the decoder and (ii) linearly interpolate between two points in the latent space. \citet{DBLP:journals/corr/MiaoYB15} apply VAE to bag-of-words representations of documents and the answer selection problem achieving good results on both tasks. \citet{DBLP:journals/corr/ZhangXS16b} and \citet{DBLP:journals/corr/SerbanSLCPCB16} apply VAE to sequence-to-sequence problems, improving over deterministic alternatives.

Various techniques to improve training of VAE models where the total cost represents a trade-off between the reconstruction cost and KL term have been used so far: KL-term annealing and input dropout~\cite{generating_sentences,ladder-vae}, imposing structured sparsity on latent variables~\cite{epitomic-vae}. In Section~\ref{subsec:optimization_difficulties} we propose another efficient technique to control the trade-off between KL and reconstruction terms.


\section{Model}

\begin{figure}
	\centering
	\includegraphics[width=0.45\textwidth]{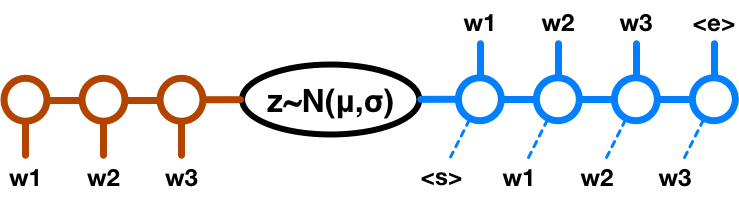}
	\caption{LSTM VAE model of \cite{generating_sentences}}
	\label{fig:lstmvae}
	\vspace{-1em}
\end{figure}

In this section we first briefly explain the VAE framework of ~\citet{vae_kingma}, then describe our hybrid architecture where the feed-forward part is composed of a fully convolutional encoder and a decoder that combines deconvolutional layers and a conventional RNN. Finally, we discuss optimization recipes that help VAE to respect latent variables, which is critical in smoothing out the latent space and being able to sample realistic sentences.

\subsection{Variational Autoencoder}
\label{subsec:vae}

The VAE is a recently introduced latent variable generative model, which combines Variational Inference with Deep Learning. It modifies the conventional Autoencoder framework in two key ways. Firstly, a deterministic internal representation \(\mathbf{z}\) (provided by the encoder) of an input \(\mathbf{x}\) is replaced with a posterior distribution \(q(\mathbf{z}|\mathbf{x})\). Inputs are then reconstructed by sampling $\mathbf{z}$ from this posterior and passing them through a decoder. To make sampling easy, the posterior distribution is usually parametrized by a Gaussian with its mean and variance predicted by the encoder. Secondly, to ensure that we can sample from any point of the latent space and still generate valid and diverse outputs, the posterior \(q(\mathbf{z}|\mathbf{x})\) is regularized with its KL divergence from a prior distribution \(p(\mathbf{z})\). The prior is typically chosen to be also a Gaussian with zero mean and unit variance, such that the KL term between posterior and prior can be computed in closed form~\cite{vae_kingma}. The total VAE cost is composed of the reconstruction term, i.e., negative log-likelihood of the data, and the KL regularizer:
\begin{equation}
\label{eq:vae_cost}
\begin{split}
J_{vae} = KL(q(\mathbf{z}|\mathbf{x})||p(\mathbf{z})) \\- \mathbb{E}_{q(\mathbf{z}|\mathbf{x})}[log\ p(\mathbf{x}|\mathbf{z})]
\end{split}
\end{equation}
\citet{vae_kingma} show that the loss function from Eq~\eqref{eq:vae_cost} can be derived from the probabilistic model perspective and it is an upper bound on the true negative likelihood of the data. 

One can view a VAE as a traditional Autoencoder with some restrictions imposed on the internal representation space. Specifically, using a sample from the \(q(\mathbf{z}|\mathbf{x})\) to reconstruct the input instead of a deterministic \(\mathbf{z}\), forces the model to map an input to a region of the space rather than to a single point. The most straight-forward way to achieve a good reconstruction error in this case is to predict a very sharp probability distribution effectively corresponding to a single point in the latent space \cite{DBLP:journals/corr/RaikoBAD14}. The additional KL term in Eq~\eqref{eq:vae_cost} prevents this behavior and forces the model to find a solution with, on one hand, low reconstruction error and, on the other, predicted posterior distributions close to the prior. Thus, the decoder part of the VAE is capable of reconstructing a sensible data sample from every point in the latent space that has non-zero probability under the prior. This allows for straightforward generation of novel samples and linear operations on the latent codes. ~\citet{generating_sentences} demonstrate that this does not work in the fully deterministic Autoencoder framework . In addition to regularizing the latent space, KL term indicaes how much information the VAE stores in the latent vector. 

\citet{generating_sentences} propose a VAE model for text generation where both encoder and decoder are LSTM networks (Figure~\ref{fig:lstmvae}). We will refer to this model as LSTM VAE in the remainder of the paper. The authors show that adapting VAEs to text generation is more challenging as the decoder tends to ignore the latent vector (KL term is close to zero) and falls back to a language model. Two training tricks are required to mitigate this issue: (i) KL-term annealing where its weight in Eq~\eqref{eq:vae_cost} gradually increases from 0 to 1 during the training; and (ii) applying dropout to the inputs of the decoder to limit its expressiveness and thereby forcing the model to rely more on the latent variables.
We will discuss these tricks in more detail in Section~\ref{subsec:optimization_difficulties}. 
Next we describe a deconvolutional layer, which is the core element of the decoder in our VAE model.

\begin{figure*}
	\centering
	\begin{tabular}{cc}
		\adjustbox{valign=b}{\subfigure[Fully feed-forward component of our VAE model\label{fig:convvae}]{%
				\includegraphics[width=.63\linewidth]{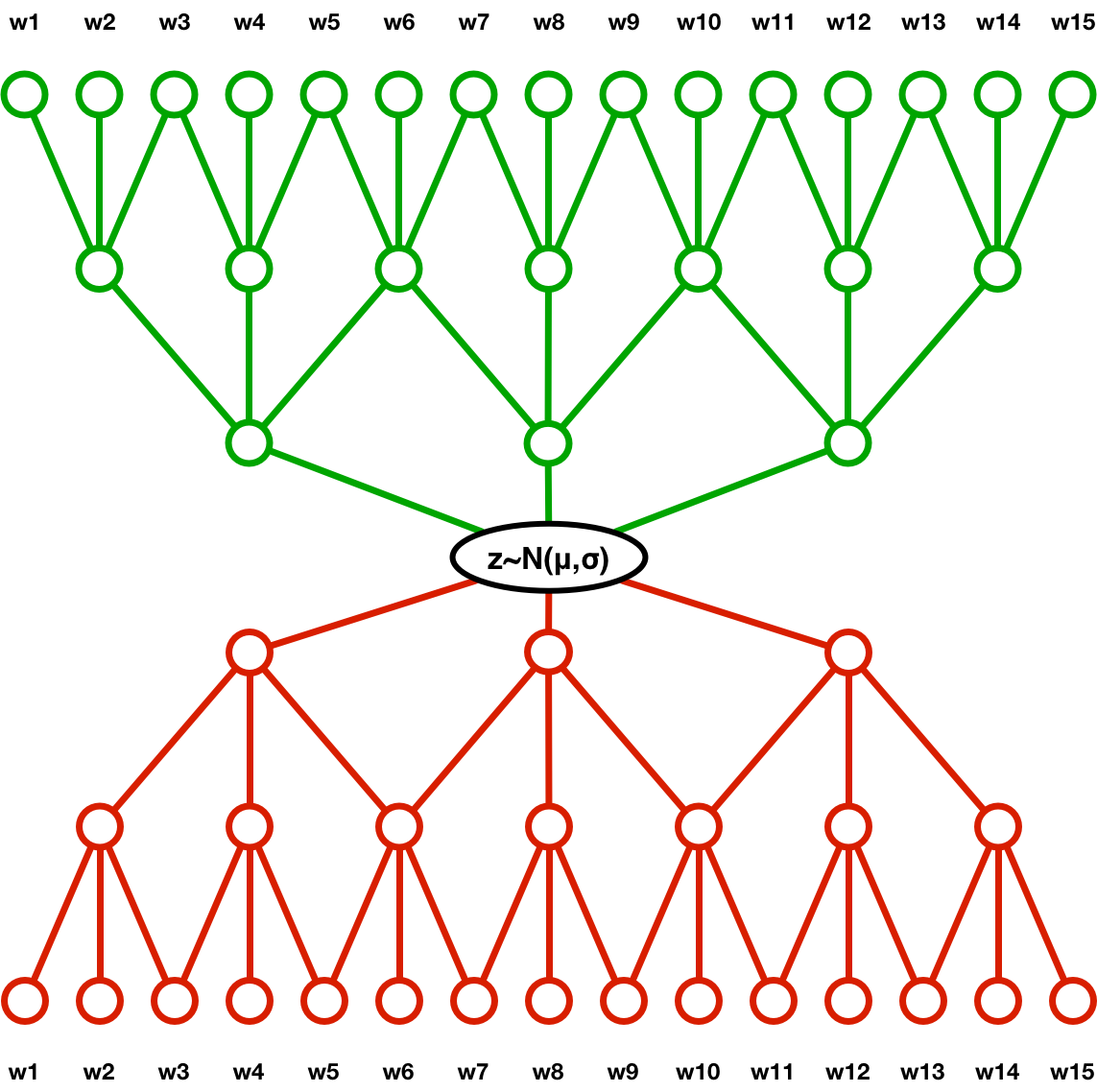}}}
		&      
		\adjustbox{valign=b}{
			\begin{tabular}{@{}c@{}}
				\subfigure[Hybrid model with LSTM decoder\label{fig:textvae_lstm}]{%
					\includegraphics[width=.27\linewidth]{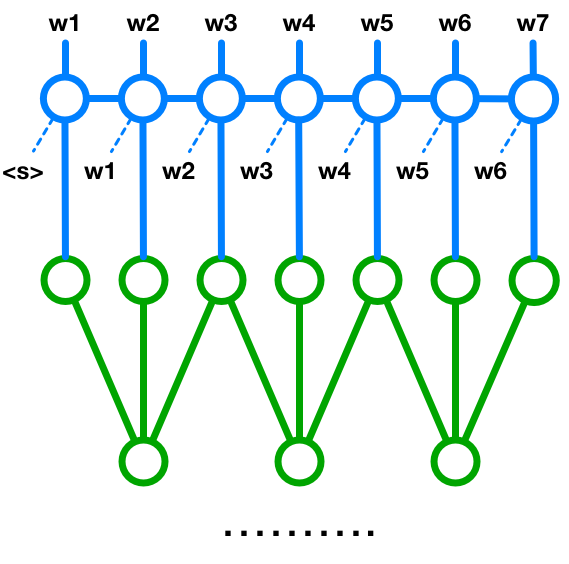}} \\
				\subfigure[Hybrid model with ByteNet decoder\label{fig:textvae_bytenet}]{%
					\includegraphics[width=.27\linewidth]{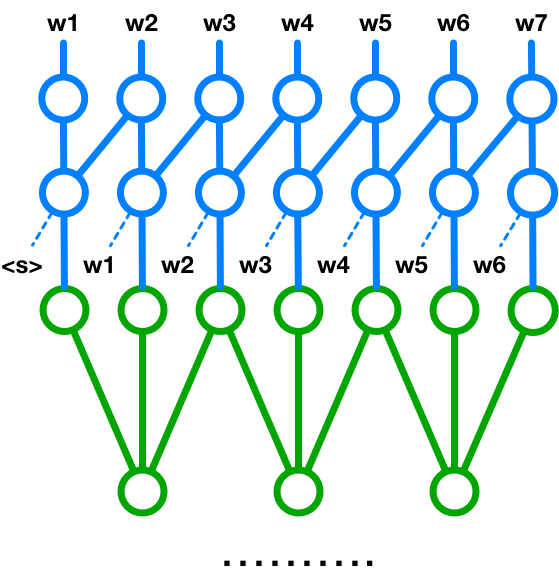}}
			\end{tabular}
		}
	\end{tabular}
	\caption{Illustrations of our proposed models.}
\end{figure*}

\subsection{Deconvolutional Networks}

A deconvolutional layer (also referred to as transposed convolutions \cite{PixelVAE:DBLP:journals/corr/GulrajaniKATVVC16} and fractionally strided convolutions \cite{DBLP:journals/corr/RadfordMC15}) performs spatial up-sampling of its inputs and is an integral part of latent variable generative models of images \cite{DBLP:journals/corr/RadfordMC15, PixelVAE:DBLP:journals/corr/GulrajaniKATVVC16} and semantic segmentation algorithms \cite{DBLP:journals/corr/NohHH15}. Its goal is to perform an ``inverse" convolution operation and increase spatial size of the input while decreasing the number of feature maps. This operation can be viewed as a backward pass of a convolutional layer and can be implemented by simply switching the forward and backward passes of the convolution operation. In the context of generative modeling based on global representations, the deconvolutions are typically used as follows: the global representation is first linearly mapped to another representation with small spatial resolution and large number of feature maps. A stack of deconvolutional layers is then applied to this representation, each layer progressively increasing spatial resolution and decreasing the amount of feature channels. The output of the last layer is an image or, in our case, a text fragment. A notable example of such a model is the Deep Network of \cite{DBLP:journals/corr/RadfordMC15} trained with adversarial objective. Our model uses a similar approach but is instead trained with the VAE objective. 

There are two primary motivations for choosing deconvolutional layers instead of the dominantly used recurrent ones: firstly, such layers have extremely efficient GPU implementations due to their fully parallel structure. Secondly, feed-forward architectures are typically easier to optimize than their recurrent counterparts, as the number of back-propagation steps is constant and potentially much smaller than in RNNs. Both points become significant as the length of the generated text increases. Next, we describe our VAE architecture that blends  deconvolutional and RNN layers in the decoder to allow for better control over the KL-term.


\subsection{Hybrid Convolutional-Recurrent VAE}
\label{subsec:convvae}
Our model is composed of two relatively independent modules. 
The first component is a standard VAE where the encoder and decoder modules are parametrized by convolutional and deconvolutional layers respectively (see Figure~\ref{fig:convvae}). This architecture is attractive for its computational efficiency and simplicity of training. 

The other component is a recurrent language model consuming activations from the deconvolutional decoder concatenated with the previous output characters. We consider two flavors of recurrent functions: a conventional LSTM network (Figure~\ref{fig:textvae_lstm}) and a stack of masked convolutions also known as the ByteNet decoder from~\citet{DBLP:journals/corr/KalchbrennerESO16} (Figure~\ref{fig:textvae_bytenet}). 
The primary reason for having a recurrent component in the decoder is to capture dependencies between elements of the text sequences -- a hard task for a fully feed-forward architecture.
Indeed, the conditional distribution $P(\mathbf{x}|\mathbf{z})=P(x_1,\dots,x_n|\mathbf{z})$ of generated sentences cannot be richly represented with a feedforward network. Instead, it is closer to: $P(x_1, \dots, x_n | \mathbf{z}) = \prod_i P(x_i | \mathbf{z})$ where components are independent of each other and are conditioned only on $\mathbf{z}$. To minimize the reconstruction cost the model is forced to encode every detail of a text fragment.
A recurrent language model instead models the full joint distribution of output sequences without  having to make independence assumptions $P(x_1,\dots,x_n|\mathbf{z})=\prod_i P(x_i|x_{i-1},\dots,x_1, \mathbf{z})$. Thus, adding a recurrent layer on top of our fully feed-forward encoder-decoder architecture relieves it from encoding every aspect of a text fragment into the latent vector and allows it to instead focus on more high-level semantic and stylistic features. 

Note that the feed-forward part of our model is different from the existing fully convolutional approaches of \citet{DBLP:journals/corr/DauphinFAG16} and \citet{DBLP:journals/corr/KalchbrennerESO16} in two respects: firstly, while being fully parallelizable during training, these models still require predictions from previous time steps during inference and thus behave as a variant of recurrent networks. In contrast, expansion of the \(z\) vector is fully parallel in our model (except for the recurrent component). Secondly, our model down- and up-samples a text fragment during processing while the existing fully convolutional decoders do not. Preserving spatial resolution can be beneficial to the overall result, but comes at a higher computational cost. 


\subsection{Optimization Difficulties}
\label{subsec:optimization_difficulties}

The addition of the recurrent component results in optimization difficulties that are  similar to those described by \citet{generating_sentences}. In most cases the model converges to a solution with a vanishingly small KL term, thus effectively falling back to a conventional language model. \citet{generating_sentences} have proposed to use input dropout and KL term annealing to encourage their model to encode meaningful representations into the \(\mathbf{z}\) vector. We found that these techniques also help our model to achieve solutions with non-zero KL term. 

KL term annealing can be viewed as a gradual transition from conventional deterministic Autoencoder to a full VAE. In this work we use linear annealing from 0 to 1. We have experimented with other schedules but did not find them to have a significant impact on the final result. As long as the KL term weight starts to grow sufficiently slowly, the exact shape and speed of its growth does not seem to affect the overall result. 

While helping to regularize the latent vector, input dropout tends to slow down convergence. We propose an alternative technique to encourage the model to compress information into the latent vector: 
in addition to the reconstruction cost computed on the outputs of the recurrent language model, we also add an auxiliary reconstruction term computed from the activations of the last deconvolutional layer:
\begin{equation}
\label{eq:aux_cost}
J_{aux} = -\mathbb{E}_{q(\mathbf{z}|\mathbf{x})}[log\ p(\mathbf{x}|\mathbf{z})].
\end{equation}
Since at this layer the model does not have access to previous output elements it needs to rely on the \(\mathbf{z}\) vector to produce a meaningful reconstruction. The final cost minimized by our model is:
\begin{equation}
\label{eq:hybrid_cost}
J_{hybrid} = J_{vae} + \alpha J_{aux}
\end{equation}
\noindent where \(\alpha\) is a hyperparameter, \(J_{aux}\) is the intermediate reconstruction term and \(J_{vae}\) is the bound from Eq~\eqref{eq:vae_cost}. 
The objective function from Eq~\eqref{eq:hybrid_cost} puts a mild constraint on the latent vector to produce features useful for historyless reconstruction. Since the autoregressive part reuses these features, it also improves the main reconstruction term. We are thus able to encode information in the latent vector without hurting expressiveness of the decoder. 

\section{Experiments}
\begin{figure*}
	\centering
	\subfigure[10 characters]{\centering\includegraphics[width=0.24\textwidth]{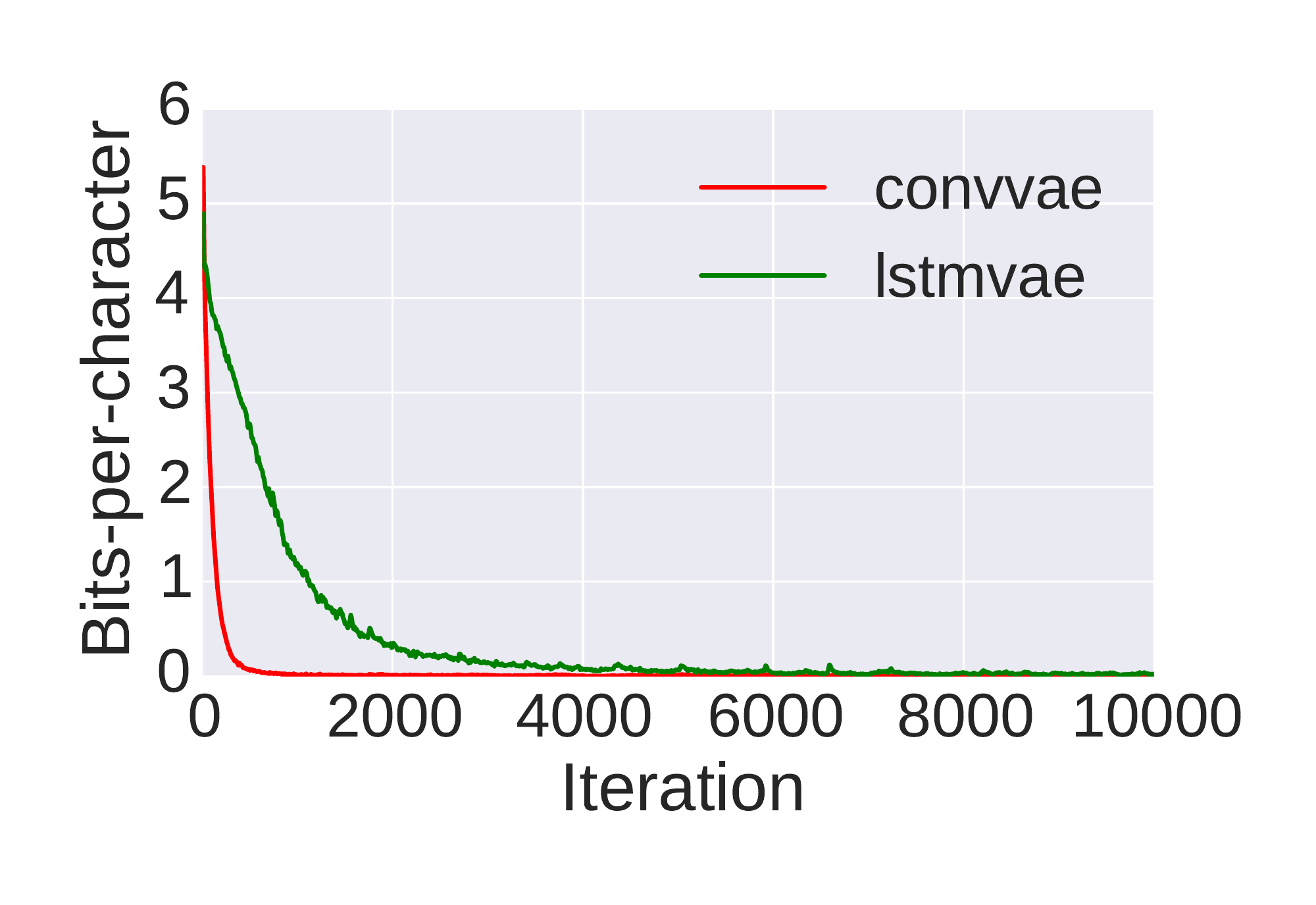}}
	\subfigure[20 characters]{\centering\includegraphics[width=0.24\textwidth]{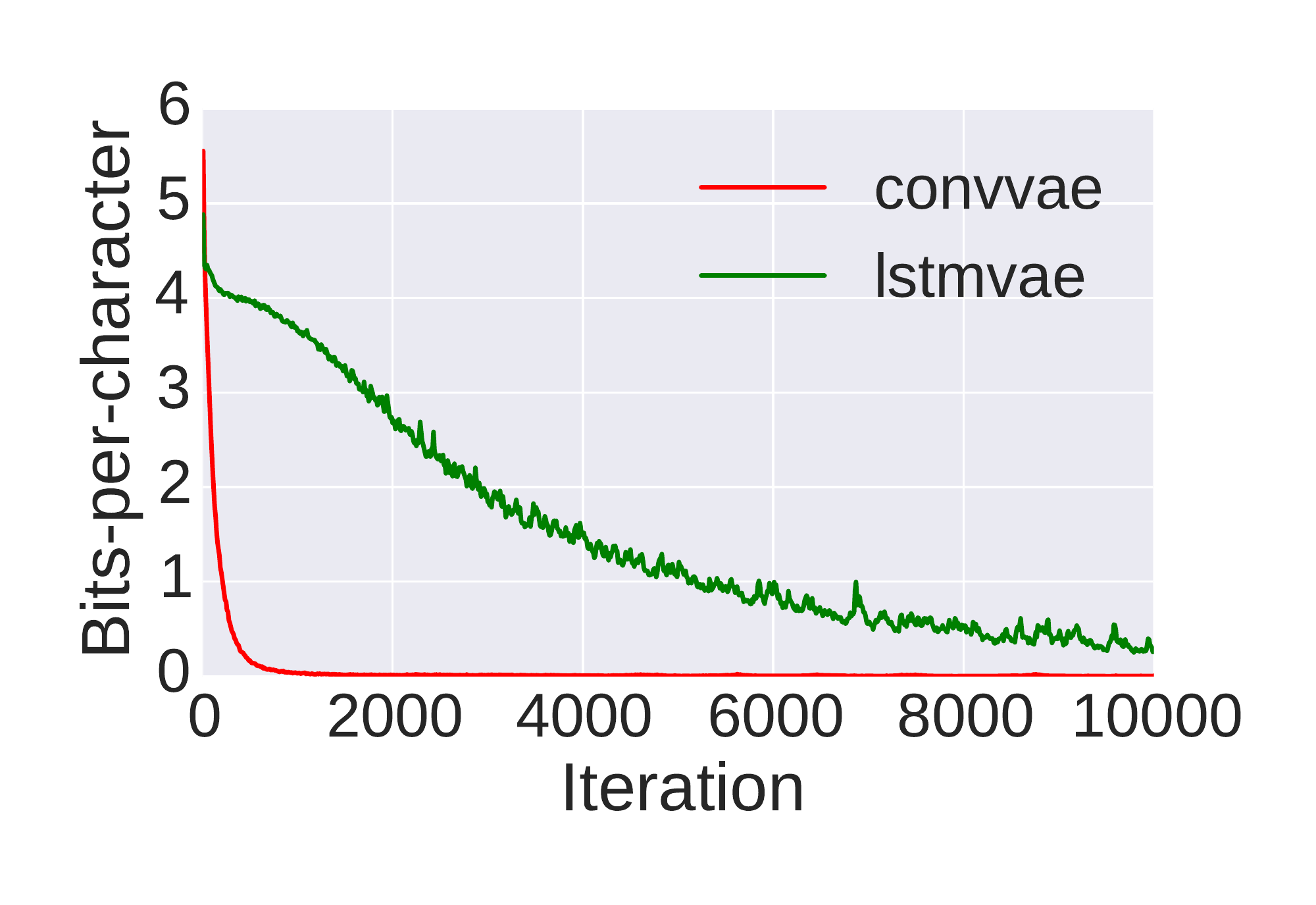}}
	\subfigure[30 characters]{\centering\includegraphics[width=0.24\textwidth]{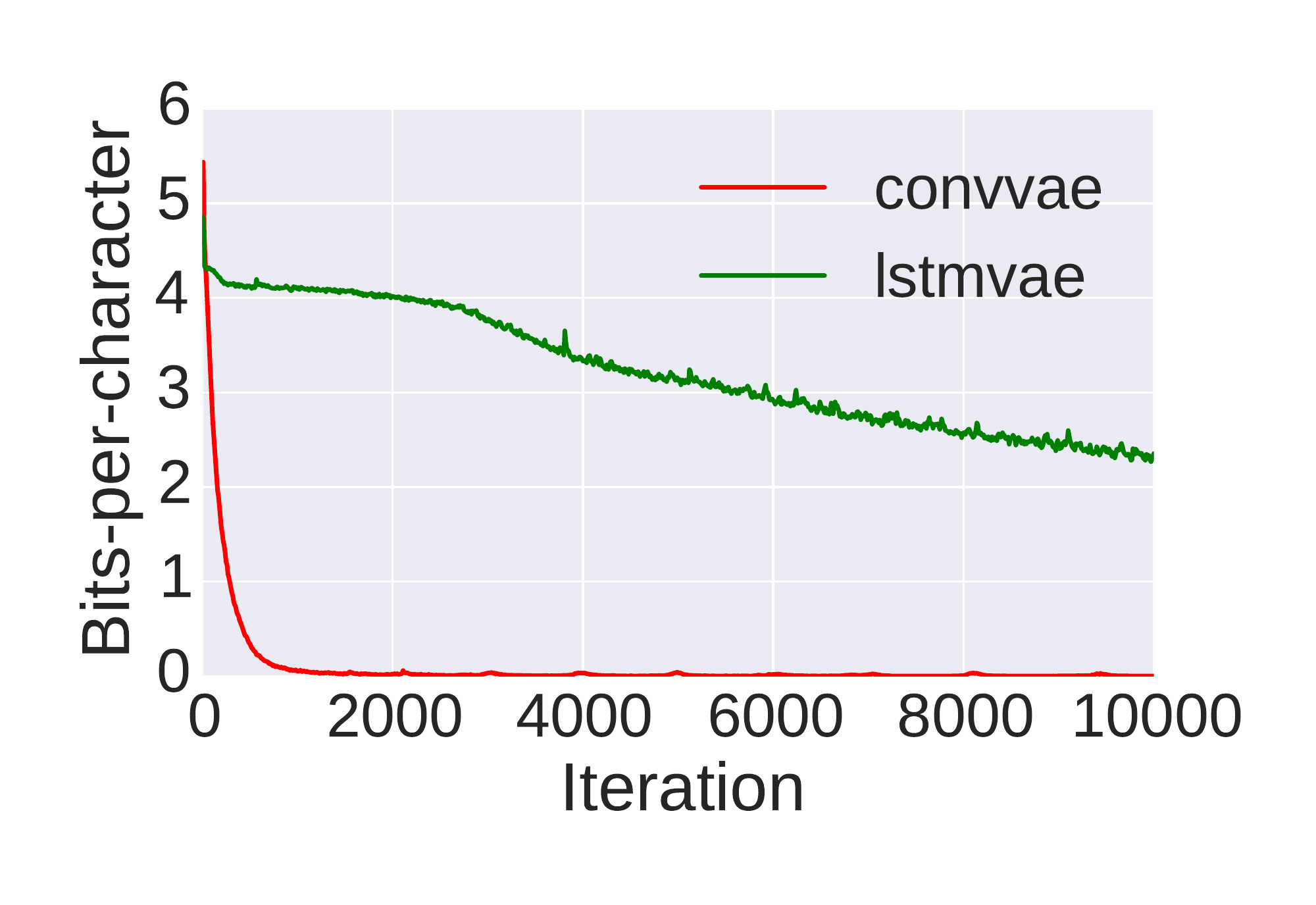}}
	\subfigure[50 characters]{\centering\includegraphics[width=0.24\textwidth]{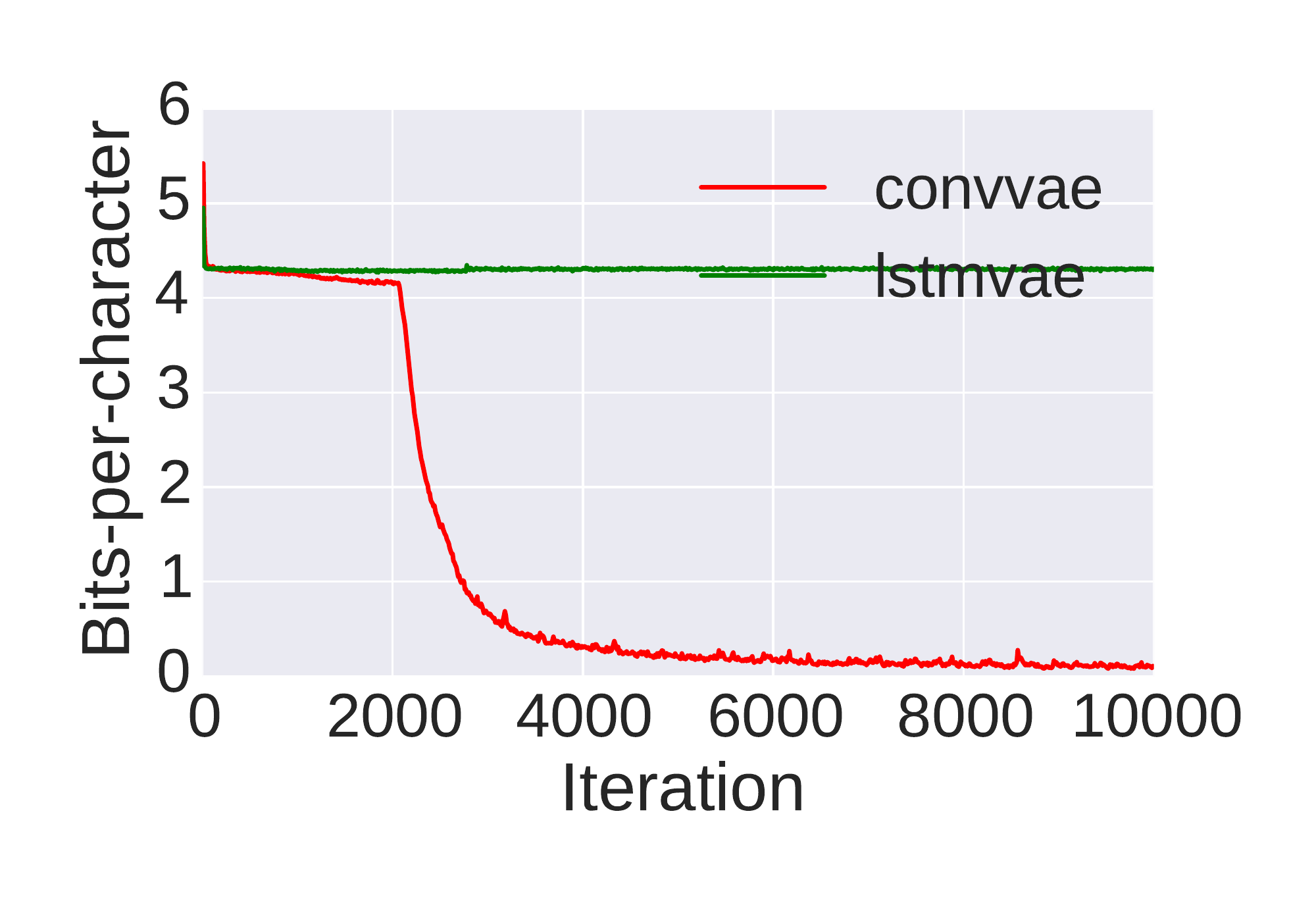}}
	\caption{Training curves of LSTM autoencoder and our model on samples of different length.}
	\label{fig:historyless}
\end{figure*}
We use KL term annealing and input dropout when training the LSTM VAE models from~\citet{generating_sentences} and KL term annealing and regularized objective function from Eq~\eqref{eq:hybrid_cost} when training our models.
All models were trained with the Adam optimization algorithm \cite{DBLP:journals/corr/KingmaB14} with decaying learning rate. We use Layer Normalization \cite{DBLP:journals/corr/BaKH16} in LSTM layers and Batch Normalization \cite{DBLP:conf/icml/IoffeS15} in convolutional and deconvolutional layers. 
To make our results easy to reproduce we have released the source code of all our experiments\footnote{\url{https://github.com/stas-semeniuta/textvae}}.

\paragraph{Data.} Our first task is character-level language generation performed on the standard Penn Treebank dataset~\cite{Marcus:1993:BLA:972470.972475}. One of the goals is to test the ability of the models to successfully learn the representations of long sequences. For training, fixed-size data samples are selected from random positions in the standard training and validation sets.

\subsection{Comparison with LSTM VAE}
\label{subsec:historylessdecoding}


\paragraph{Historyless decoding.} We start with an experiment where the decoder is forced to ignore the history and has to rely fully on the latent vector. By conditioning the decoder only on the latent vector \(\mathbf{z}\) we can directly compare the expressiveness of the compared models.
For the LSTM VAE model historyless decoding is achieved by using the dropout on the input elements with the dropout rate equal to 1. We compare it to our fully-feedforward model without the recurrent layer in the decoder (Figure~\ref{fig:convvae}). Both networks are parametrized to have equal number of parameters.

To test how well both models can cope with the stochasticity of the latent vectors, we minimize only the reconstruction term from Eq.~\eqref{eq:vae_cost}. This is equivalent to a pure Autoencoder setting with stochastic internal representation and no regularization of the latent space. This experiment corresponds to an initial stage of training with KL term annealing when its weight is set to 0.

The results are presented in Figure~\ref{fig:historyless}. Note that when the length of input samples reaches 30 characters, the historyless LSTM autoencoder fails to fit the data well, while the convolutional architecture converges almost instantaneously. The results appear even worse for LSTMs on sequences of 50 characters.
To make sure that this effect is not caused by optimization difficulties, i.e. exploding gradients \cite{DBLP:conf/icml/PascanuMB13}, we have searched over learning rates, gradient clipping thresholds and sizes of LSTM layers but were only able to get results comparable to those shown in Figure~\ref{fig:historyless}. Note that LSTM networks make use of Layer Normalization \cite{DBLP:journals/corr/BaKH16} which has been shown to make training of such networks easier. These results suggest that our model is easier to train than the LSTM-based model, especially for modeling longer pieces of text. Additionally, our model is computationally faster by a factor of roughly two, since we run only one recurrent network per sample and time complexity of the convolutional part is negligible in comparison.


\begin{figure}
	\centering
	\includegraphics[height=0.35\textwidth]{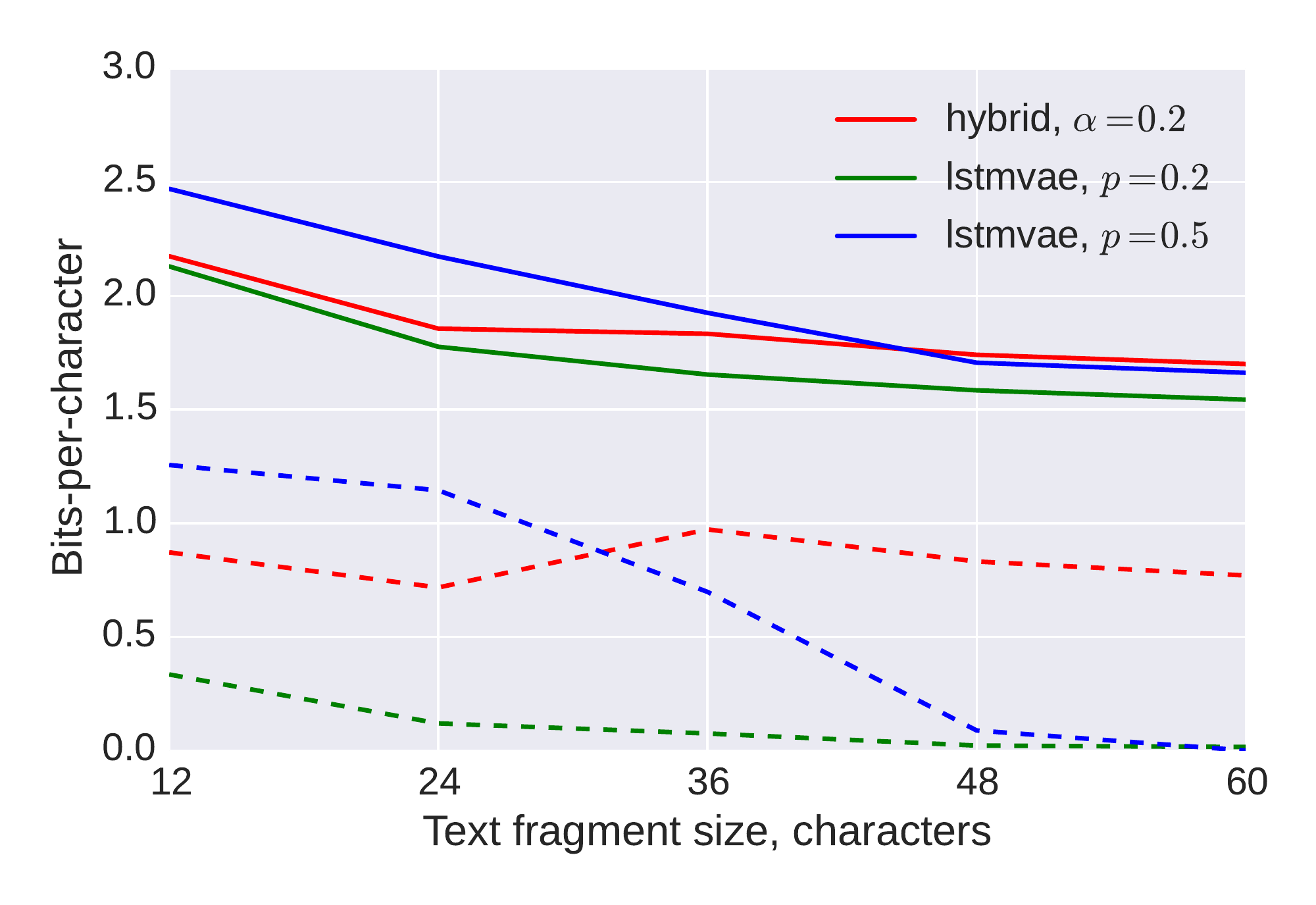}
	\caption{The full cost (solid lines) and its KL component (dashed lines) in bits-per-character of our Hybrid model trained with 0.2 \(\alpha\) hyper-parameter vs. LSTM based VAE trained with 0.2 and 0.5 input dropout.}
	\label{fig:ratios}
	\vspace{-1em}
\end{figure}

\paragraph{Decoding with history.} We now move to a case where the decoder is conditioned on both the latent vector and previous output elements. In these experiments we pursue two goals: firstly, we verify whether the results obtained on the historyless decoding task also generalize to a less restricted case. Secondly, we study how well the models cope with stochasticity introduced by the latent variables.

We fix input dropout rates at 0.2 and 0.5 for LSTM VAE and use auxiliary reconstruction loss (Section~\ref{subsec:optimization_difficulties}) with 0.2 weight in our Hybrid model. The bits-per-character scores on differently sized text samples are presented in Figure~\ref{fig:ratios}. As discussed in Section~\ref{subsec:vae}, the KL term value indicates how much information the network stores in the latent vector.
We observe that the amount of information stored in the latent vector by our model and the LSTM VAE is comparable when we train on short samples and largely depends on hyper-parameters \(\alpha\) and \(p\) .
When the length of a text fragment increases, LSTM VAE is able to put less information into the latent vector (i.e., the KL component is small) and for texts longer than 48 characters, the KL term drops to almost zero while for our model the ratio between KL and reconstruction terms stays roughly constant. This suggests that our model is better at encoding latent representations of long texts since the amount of information in the latent vector does not decrease as the length of a text fragment grows. In contrast, there is a steady decline of the KL term of the LSTM VAE model. This result is consistent with our findings from the historyless decoding experiment. Note that in both of these experiments the LSTM VAE model fails to produce meaningful latent vectors with inputs over 50 characters long. This further suggests that our Hybrid model encodes long texts better than the LSTM VAE.

\subsection{Controlling the KL term}

\begin{figure}
	\centering
	\includegraphics[height=0.35\textwidth]{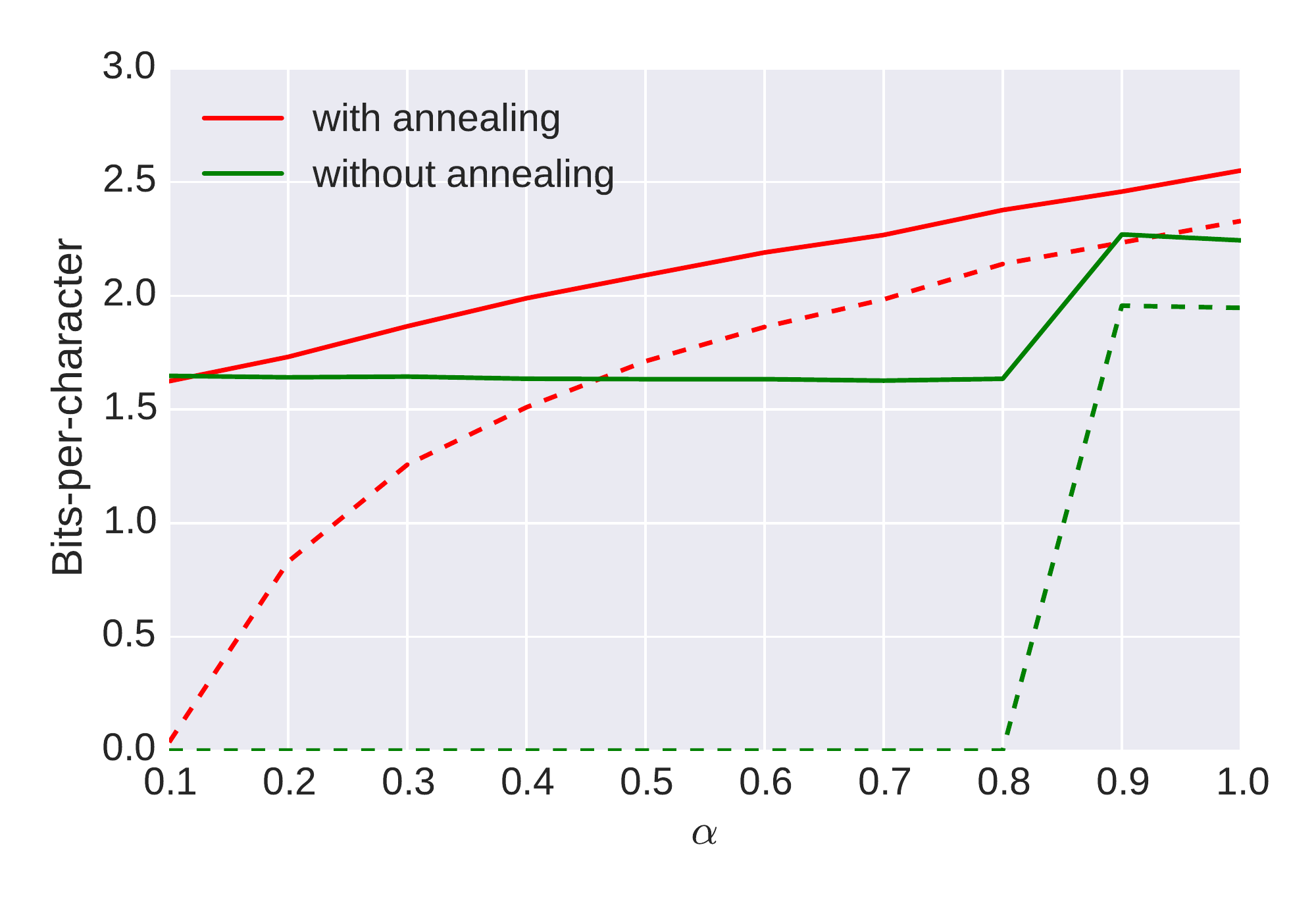}
	\caption{The full cost (solid line) and the KL component (dashed line) of our Hybrid model with LSTM decoder trained with various \(\alpha\), with and without KL term weight annealing.}
	\label{fig:alpha}
\end{figure}

We study the effect of various training techniques that help control the KL term which is crucial for training a generative VAE model.

\paragraph{Aux cost weight.} First, we provide a detailed view of how optimization tricks discussed in Section~\ref{subsec:optimization_difficulties} affect the performance of our Hybrid model. Figure \ref{fig:alpha} presents results of our model trained with different values of \(\alpha\) from Eq.~\eqref{eq:hybrid_cost}. Note that the inclusion of the auxiliary reconstruction loss slightly harms the bound on the likelihood of the data but helps the model to rely more on the latent vector as \(\alpha\) grows. A similar effect on model's bound was observed by \citet{generating_sentences}: increased input dropout rates force their model to put more information into the \(\mathbf{z}\) vector but at the cost of increased final loss values. This is a trade-off that allows for sampling outputs in the VAE framework.
Note that our model can find a solution with non-trivial latent vectors when trained with the full VAE loss provided that the \(\alpha\) hyper-parameter is large enough. Combining it with KL term annealing helps to find non-zero KL term solutions at smaller $\alpha$ values.

\begin{figure}
	\centering
	\includegraphics[height=0.35\textwidth]{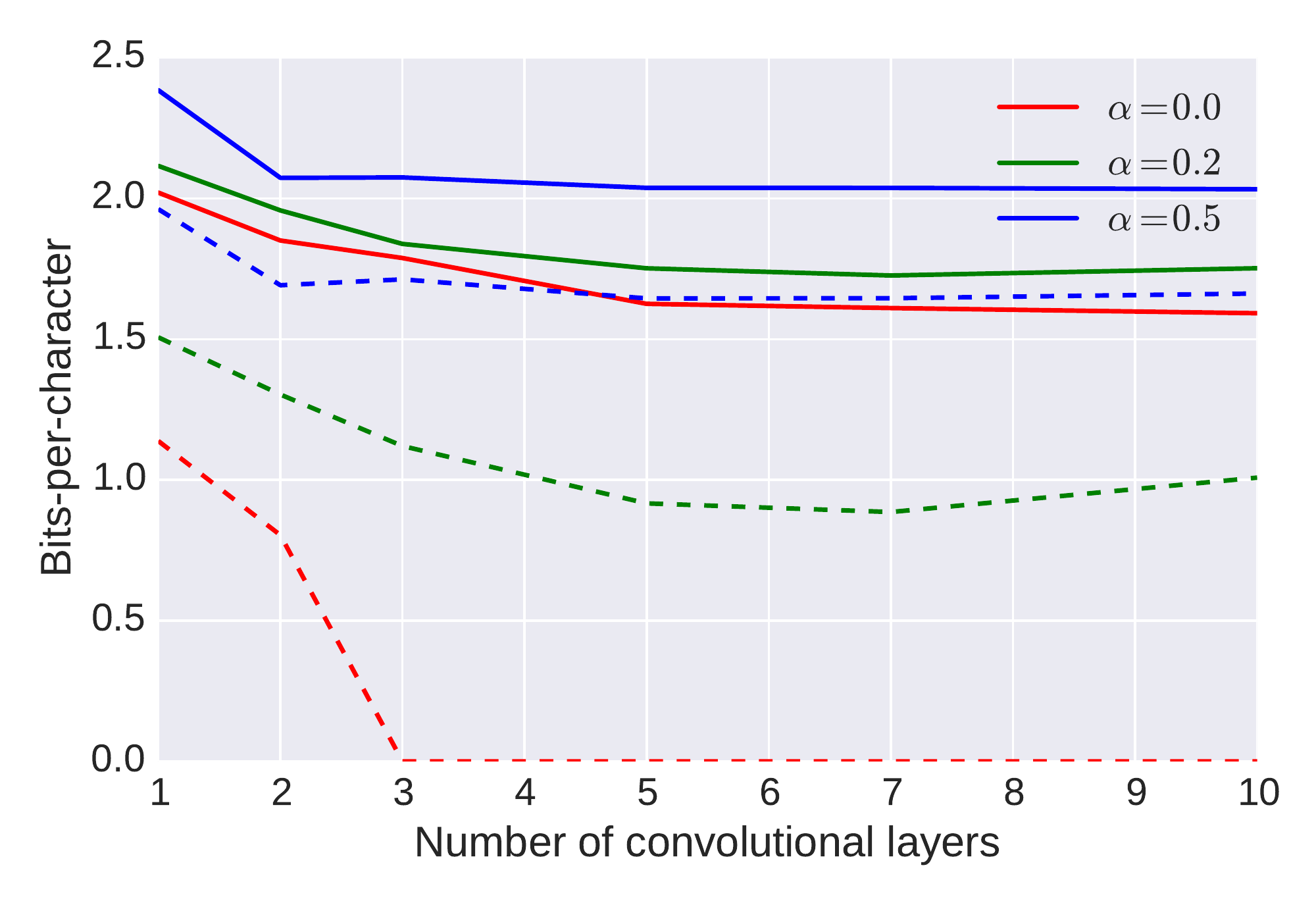}
	\caption{The full cost (solid line) and the KL component (dashed line) of our Hybrid model with ByteNet decoder trained with various number of convolutional layers.}
	\label{fig:bytenet_decoders}
\end{figure}

\begin{table*}
	\footnotesize
	\begin{tabular}{l}
		\toprule
		@userid @userid @userid @userid @userid ...                                       \\
		I want to see you so much  @userid  \#FollowMeCam ...                             \\
		@userid @userid @userid @userid @userid ...                                       \\
		Why do I start the day today?                                                     \\
		 \midrule
		@userid thanks for the follow back                                                \\
		no matter what I'm doing with my friends  they are so cute                        \\
		@userid and I have to do that for a couple of days and then I can start with them \\
		I wanna go to the UK tomorrow!!  \#feelinggood \#selfie \#instago                 \\
		@userid @userid I'll come to the same time and it was a good day too  xx          \\ \bottomrule
	\end{tabular}
	\caption{Random sample tweets generated by LSTM VAE (top) and our Hybrid model (bottom).}
	\label{table:twitter_samples}
\end{table*}

\paragraph{Receptive field.} The goal of this experiment is to study the relationship between the KL term values and the expressiveness of the decoder. Without KL term annealing and input dropout, the RNN decoder in LSTM VAE tends to completely ignore information stored in the latent vector and essentially falls back to an RNN language model.
To have a full control over the receptive field size of the recurrent component in our decoder, we experiment with masked convolutions (Figure~\ref{fig:textvae_bytenet}), which is similar to the decoder in ByteNet model from~\citet{DBLP:journals/corr/KalchbrennerESO16}. We fix the size of the convolutional kernels to 2 and do not use dilated convolutions and skip connections as in the original ByteNet. 

The resulting receptive field size of the recurrent layer in our decoder is equal to \(N+1\) characters, where \(N\) is the number of convolutional layers. We vary the number of layers to find the amount of preceding characters that our model can consume without collapsing the KL term to zero. 

Results of these experiments are presented in Figure~\ref{fig:bytenet_decoders}. Interestingly, with the receptive field size larger than 3 and without the auxiliary reconstruction term from Eq.~\eqref{eq:hybrid_cost} ($\alpha=0$) the KL term collapses to zero and the model falls back to a pure language model. This suggests that the training signal received from the previous characters is much stronger than that from the input to be reconstructed. Using the auxiliary reconstruction term, however, helps to find solutions with non-zero KL term component irrespective of receptive field size. Note that increasing the value of \(\alpha\) results in stronger values of KL component. This is consistent with the results obtained with LSTM decoder in Figure~\ref{fig:alpha}.

\subsection{Generating Tweets}
\label{section:tweets}
In this section we present qualitative results on the task of generating tweets.

\paragraph{Data.} We use 1M tweets\footnote{a random sample collected using the Twitter API} to train our model and test it on a held out dataset of 10k samples. We minimally preprocess tweets by only replacing user ids and urls with ``@userid" and ``url". 

\paragraph{Setup.} We use 5 convolutional layers with the ReLU non-linearity, kernel size 3 and stride 2 in the encoder. The number of feature maps is [128, 256, 512, 512, 512] for each layer respectively.  The decoder is configured equivalently but with the amount of feature maps decreasing in each consecutive layer. The top layer is an LSTM with 1000 units. We have not observed significant overfitting. The baseline LSTM VAE model contained two distinct LSTMs both with 1000 cells. The models have comparable number of parameters: 10.5M for the LSTM VAE model and 10.8M for our hybrid model.

\begin{table}
	\begin{center}
		\begin{tabular}{r|r|r}
			                       &  Rec &   KL \\ \bottomrule
			   LSTM VAE, \(p=0.2\) & 67.4 &  1.0 \\
			   LSTM VAE, \(p=0.5\) & 77.1 &  2.1 \\
			   LSTM VAE, \(p=0.8\) & 93.7 &  3.8 \\
			Hybrid, \(\alpha=0.2\) & 58.5 & 12.5 \\ \bottomrule
		\end{tabular}
	\end{center}
	\caption{Breakdown into KL and reconstruction terms for char-level tweet generation. \(p\) refers to input dropout rate.}
	\label{table:charlevel_tweet_results}
	\vspace{-1em}
\end{table}


\paragraph{Results.} Both VAE models are trained on the character-level generation. The breakdown of total cost into KL and reconstruction terms is given in Table~\ref{table:charlevel_tweet_results}. Note that while the total cost values are comparable, our model puts more information into the latent vector, further supporting our observations from Section~\ref{subsec:historylessdecoding}. This is reflected in the random samples from both models, presented in Table~\ref{table:twitter_samples}. We perform greedy decoding during generation so any variation in samples is only due to the latent vector. LSTM VAE produces very limited range of tweets and tends to repeat "@userid" sequence, while our model produces much more diverse samples.
\section{Conclusions}

We have introduced a novel generative model of natural texts based on the VAE framework. Its core components are a convolutional encoder and a deconvolutional decoder combined with a recurrent layer. We have shown that the feed-forward part of our model architecture makes it easier to train a VAE and avoid the problem of KL-term collapsing to zero, where the decoder falls back to a standard language model thus inhibiting the sampling ability of VAE. 
Additionally, we propose a more natural way to encourage the model to rely on the latent vector by introducing an additional cost term in the training objective. We observe that it works well on long sequences which is hard to achieve with purely RNN-based VAEs using the previously proposed tricks such as KL-term annealing and input dropout. Finally, we have extensively evaluated the trade-off between the KL-term and the reconstruction loss. In particular, we investigated the effect of the receptive field size on the ability of the model to respect the latent vector which is crucial for being able to generate realistic and diverse samples.
In future work we plan to apply our VAE model to semi-supervised NLP tasks and experiment with conditioning generation on various text attributes such as sentiment and writing style. 


\section*{Acknowledgments}

We thank Enrique Alfonseca, Katja Filippova, Sylvain Gelly, Jason Lee and David Weiss for their useful feedback while preparing this draft.
This project has received funding from the European Union's Framework Programme for Research and Innovation HORIZON 2020 (2014-2020) under the Marie Skłodowska-Curie Agreement No. 641805. Stanislau Semeniuta thanks the support from Pattern Recognition Company GmbH. We gratefully acknowledge the support of NVIDIA Corporation with the donation of the Titan X GPU used for this research.

\bibliography{acl2017}

\appendix
\section{Supplementary Material}
In this supplemental material we include more random samples from our hybrid VAE model and LSTM VAE from~\cite{generating_sentences}. We also perform linear operations in the latent space to demonstrate that they result in smooth and syntactically correct transitions between generated texts.


\subsection{Random Samples}

Table~\ref{table:charvae_samples} contains random samples generated by our hybrid model with LSTM decoder while Table~\ref{table:lstmvae_samples} contains samples from LSTM VAE. As already demonstrated by our experiments in Section~\ref{section:tweets}, for LSTM VAE model it is much harder to find solutions with KL terms significantly different from zero, which results in redundant and uninteresting samples generated by the model. We have experimented with larger dropout rates but found that this did not lead to better samples. In contrast, samples from our model have much more variation both in the length and the content. 

\begin{table*}
	\footnotesize
	\begin{tabular}{l}
		@userid you're welcome  hope you had a good day                                                                    \\
		@userid you deserve it!!!                                                                                          \\
		Good night twitter fam                                                                                             \\
		I love this place to be on @userid  what a good morning  \#happybirthday \#happy url                               \\
		@userid love you too  @userid                                                                                      \\
		My sister is the best thing to do with me                                                                          \\
		@userid I LOVE YOU TOO                                                                                             \\
		Please look @userid been.                                                                                          \\
		@userid you're welcome  i love you so much \&lt;3 \&lt;3 \&lt;3                                                    \\
		@userid lol sorry for the show  I am so excited for tomorrow  what a good day to see it with you all to see you  x \\
		Please let me know what to do                                                                                      \\
		@userid @userid @userid well done  I want to see you again so we can find a show tomorrow  \#happybirthday         \\
		@userid no problem  ... I will be fine thank you for the follow                                                    \\
		@userid Hi ! Can you rt this tweet please ? this is very important to me  I love you :3 photo by @userid  2        \\
		Now that is my birthday and the weekend is over!!                                                                  \\
		If you want the song that you can't wait                                                                           \\
		@userid  hey could you please vote for my band Sait Believe to open for I am giant ?  url                          \\
		@userid I am so thank you for the shout out there! Happy to help out with your tweet                               \\
		@userid they said my fave would notice me this month, will you?  xx i love you xx400                               \\
		@userid @userid  hope the second season                                                                            \\
		@userid Hello   How are you doing?                                                                                 \\
		And I wanna be the same town                                                                                       \\
		@userid I know  haha xx                                                                                            \\
		@userid aww  thanks so much for the follow!  have a great day!  \#CornettoWelcomesTaylorToManila                   \\
		@userid ... I love you so much  I love you so much \&lt;3 \&lt;3                                                   \\
		Well I don't have to wait till the 19th  url                                                                       \\
		Goodnight friends\\
		@userid for that person and they are all good thank you for the shoutout  \#SelenaForMMVA \\
		@userid Hello mate! We're glad you like it \\
		"@Sarah\_Bear: Why not?  \#SelenaForMMVA \\
		When i see the start to the fault in our \\
		\#tbt to the first 10 mins of the day I have to wait till 10 \\
	 	Looking forward for @userid at \#TFIOS tonight with the graduation tomorrow!  \#GoodMorning \#StarCrossed \\
		@userid  Hi can. I love you so much \&lt;3 \\
		@userid I will be on FB and we can try it \\
		@userid hey cam, i love you so much, follow me please, make me happy cam  \#FollowMeCam \#FollowMeCam \#FollowMeCam 116 \\
		I'm gonna miss my best friends  \#happy \#instafood \#love url \\
		Don't forget to do more than a month... url \\
		@userid like that... I have a lot of time to take a look  \#startup \#followmecam x \\
		@userid you should get that that would be a great place \\
		
	\end{tabular}
	\caption{Random samples from ou hybrid model trained with \(\alpha=0.2\).}
	\label{table:charvae_samples}
\end{table*}

\begin{table*}
	\footnotesize
	\begin{tabular}{l}
		@userid @userid @userid @userid @userid @userid @userid @userid @userid ...      \\
		@userid thank you                                                                \\
		I want to see you so much  please follow me  \#FollowMeCam @userid  x1           \\
		Lood morning  @userid  url                                                       \\
		@userid @userid @userid @userid @userid @userid @userid @userid @userid ...      \\
		I want to see you so much  please pick me @userid PLEASE  \#KedsRedTour ...      \\
		I love the best to me the best on the world to me                                \\
		@userid @userid @userid @userid @userid @userid @userid @userid @userid ...      \\
		I want to see you to me that the world to me                                     \\
		@userid @userid @userid @userid @userid @userid @userid @userid @userid ...      \\
		@userid @userid @userid @userid @userid @userid @userid @userid @userid ...      \\
		@userid thank you  xx                                                            \\
		I love you so much @userid   @userid     xxx                                     \\
		I want to see you so much @userid   @userid @userid @userid @userid @userid  x15 \\
		@userid @userid @userid @userid @userid @userid @userid @userid @userid ...      \\
		I want to see you so much @userid   I love you so much  x  \#FollowMeCam ...
	\end{tabular}

	\caption{Random samples from the LSTM VAE model with \(p=0.5\). We have truncated samples that repeat the same hashtag or "@userid" sequence.}
	\label{table:lstmvae_samples}
\end{table*}


\subsection{Linear operations in the latent space}

\citet{generating_sentences} have shown that LSTM-based VAE for text produces latent representations such that linear operations on latent vectors result in meaningful transformations of generated texts. 
In Table~\ref{table:interpolations} we present interpolation examples produced by our hybrid model. Note gradual changes in length of generated tweets, consistent usage of topics along trajectories and generally sensible structure.

\begin{table*}
	\footnotesize
	\begin{tabular}{l}
		\textbf{@userid I do that too. I have to wait a lot of people too}                                            \\
		@userid I do not know about it and then I can find a contest                                                  \\
		@userid I am so excited for this summer  I hope you are well                                                  \\
		@userid i don't know what to do in the morning  i love you so much for the shoutout                           \\
		@userid i don't know what to do if you don't mind follow me  i love you so much xx                            \\
		@userid it would be awesome to hear  it's a great place to see you around the weekends                        \\
		\textbf{@userid it would be awesome to hear  I'm so excited for the summer  I'm going to see them}            \\ \midrule
		\textbf{Just been to my best friend and it was a great time in the world and I have to wait till the weekend} \\
		Just been to my best friend and it was a great time with the story of the day  \#happy \#goodnight            \\
		Just been to my best friend and it was a great time with the stream  \#happy \#summer \#sunshine              \\
		So the first of my students are all going to the start to the start of the day  \#happy \#goodnight           \\
		So the fault in our stars is the best day of the week and it's all good                                       \\
		@userid LOL I'm just going to see you and the students to the start                                           \\
		\textbf{@userid LOL I'm just going to see you at the end of the day  hahaha}                                  \\ \midrule
		\textbf{@userid @userid I will have to wait till I'm off to school tomorrow  i hope you get a good day  xx}   \\
		@userid @userid I will have to wait till I'm done with you and the world is amazing  how are you?             \\
		@userid @userid I will have to see it tomorrow for the first time with you and the best for you               \\
		@userid So sad  I have to go back to the pool but it's a beautiful day for the weekend                        \\
		@userid  I am so excited for you to come to the pool and I will see it tomorrow                               \\
		There's a few more weeks of summer and the world will be @userid                                              \\
		\textbf{Make it a good day today  \#summer \#selfie \#girls \#selfie @userid}                                 \\ \midrule
		\textbf{@userid  I am so sorry to hear that}                                                                  \\
		@userid  I am so sorry                                                                                        \\
		@userid  I was a good day                                                                                     \\
		@userid  I want to see it                                                                                     \\
		@userid aw thank you                                                                                          \\
		@userid my pleasure                                                                                           \\
		\textbf{@userid my pleasure}
	\end{tabular}
	
	\caption{Linear interpolations between two randomly selected points in the latent space generated by our hybrid model. Sentences representing begin and end points are bolded.}
	\label{table:interpolations}
\end{table*}

\end{document}